# TRAFFIC ACCIDENT ANALYSIS USING DECISION TREES AND NEURAL NETWORKS

Miao M. Chong, Ajith Abraham, Marcin Paprzycki
*Computer Science Department, Oklahoma State University, USA*
`{miaoc,marcin,aa}@cs.okstate.edu`

## ABSTRACT

The costs of fatalities and injuries due to traffic accident have a great impact on society. This paper presents our research to model the severity of injury resulting from traffic accidents using artificial neural networks and decision trees. We have applied them to an actual data set obtained from the National Automotive Sampling System (NASS) General Estimates System (GES). Experiment results reveal that in all the cases the decision tree outperforms the neural network. Our research analysis also shows that the three most important factors in fatal injury are: driver's seat belt usage, light condition of the roadway, and driver's alcohol usage.

## 1. INTRODUCTION AND RELATED RESEARCH

In recent years, researchers have been utilizing real-life data in studying various aspects of driver injuries resulting from traffic accidents. Their research can be divided into three general areas namely prediction of the injury severity, establishing the most important factors influencing the severity of the injury and miscellaneous projects related to traffic accidents. Related important works can be summarized as follows.

Abdelwahab et al. [2] studied the 1997 accident data for the Central Florida area focusing on two-vehicle accidents that occurred at signalized intersections. The injury severity was divided into three classes: no injury, possible injury and disabling injury. The performance of Neural Network (NN) trained by Levenberg-Marquardt algorithm and Fuzzy ARTMAP were compared, and found that NN (65.6% and 60.4% classification accuracy for the training and testing phases) performed better than Fuzzy ARTMAP (56.1%).

Shankar, et al. [14] applied a nested logic formulation for estimating accident severity likelihood conditioned on the occurrence of an accident. The study found that there is a greater probability of evident injury or disabling injury/fatality relative to no evident injury if at least one driver did not use a restraint system at the time of the accident. Kim et al. [8] developed a log-linear model to clarify the role of driver characteristics and behaviors in the causal sequence leading to more severe injuries. They found that driver behaviors of alcohol or drug use and lack of seat belt use greatly increase the odds of more severe crashes and injuries. Bedard et al. [3] applied a multivariate logistic regression to determine the independent contribution of driver, crash, and vehicle characteristics to drivers' fatality risk. It was found that increasing seatbelt use, reducing speed, and reducing the number and severity of driver-side impacts might prevent fatalities. Yang, et al. used NN approach to detect safer driving patterns that have less chances of causing death and injury when a car crash occurs [17]. They performed the Cramer's V Coefficient test [18] to identify significant variables that cause injury, therefore, reduced the dimensions of the data for the analysis. The 1997 Alabama interstate alcohol-related data was used and was found that by controlling a single variable (such as driving speed, or light conditions) fatalities and injuries could be reduced by up to 40%. Ossiander et al. [13] used Poisson regression to analyze the association between the fatal crash rate (fatal crashes per vehicle mile traveled) and the speed limit increase [13] and found that the speed limit increase was associated with a higher fatal crash rate and more deaths on freeways in Washington State. Some researchers studied the relationship between drivers' age, gender, vehicle mass, impact speed or driving speed measure with fatalities [4, 9, 10, 11, 16].

Mussone et al. [12] used NN to analyze vehicle accident that occurred at intersections in Milan, Italy. Results showed that the highest accident index for running over of pedestrian occurs at non-signalized intersections at nighttime. Dia et al. used real-world data for developing a multi-layered NN freeway incident detection model [5]. Results showed that NN could provide faster and more reliable incident detection over the model that was in operation on Melbourne's freeways. Abdel-Aty et al. [1] used the Fatality Analysis Reporting System (FARS) crash databases covering the period of 1975-2000 to analyze the effect of the increasing number of light truck vehicle (LTV) registrations on fatal angle collision trends in the US. They investigated the number of annual fatalities that result from angle collisions as well as collision configuration (car-car, car-LTV, LTV-car, and LTV-LTV). Modeling results showed that fatalities in angle collisions will increase in the next 10 years, and that they are affected by the expected increase in the percentage of LTVs in traffic. Evanco conducted a multivariate population-based statistical analysis to determine the relationship between fatalities and accident notification times [6]. The analysis demonstrated that accident notification time is an important determinant of the number of fatalities for accidents on rural roadways. Sohn et al. applied data fusion, ensemble and clustering to improve the accuracy of individual classifiers for two categories of severity (bodily injury and property damage) of road traffic accident [15]. They applied a clustering algorithm to the dataset to divide the data into subsets of data, and then used each subset of data to train the classifiers (NN and decision trees). The studies revealed that the classification based on clustering works better if the variation in observations is relatively large as in Korean road traffic accident data.

The main goal of our work was similar to that presented in [2]. Using 'real world' accident data we have utilized known data mining techniques to develop a realistic model of injury severity resulting from an automobile accident. Our goals were to improve the accuracy of prediction over that reported in [2], and to establish the most important factors influencing the severity of the injury. In this paper we present a report on our attempt at utilizing decision trees and neural networks. The remaining parts of the paper are organized as follows. In Section 2, we discuss the accident data set used in our work. Section 3 is devoted to introduction of the machine learning techniques utilized in our work. Performance results and their analysis are presented in Section 4.

## 2. ACCIDENT DATA SET

This study used dataset from the National Automotive Sampling System (NASS) General Estimates System (GES). The GES data are intended to be a nationally representative probability sample from the annual estimated 6.4 million police accident reports in the USA. The initial dataset for the study contains traffic accident records from 1995 to 2000, a total number of 417,670 cases. It consists of label-variables: *year*, *month, region*, *primary sampling unit*, *the number of the police jurisdiction*, *case number, person number, vehicle number, vehicle make and model;* 'information carrying' variables that are to be used as inputs: *drivers' age*, *gender, alcohol usage, restraint system, eject, vehicle body type, vehicle age, vehicle role, initial point of impact, manner of collision, rollover, roadway surface condition, light condition, travel speed, speed limit* and the accident result variable that is to be used as an output: *injury severity*. As reported in our dataset, five label-values represent the injury severity: *no injury, possible injury, non-incapacitating injury, incapacitating injury*, and *fatal injury*.

Since the head-on collision has the highest percent of fatal injury records, we have narrowed the dataset down to head-on collision only. As a result we have ended with a total of 10,386 records. Among these, there were 160 records of head-on collision with fatal injury and all of these records have the impact point categorized as *front impact*. This prompted us to analyze the data further and established that the initial point of impact has 9 categories: *no damage/non-collision, front, right side, left side, back, front right corner, front left corner, back right corner,* and *back left corner*. The head-on collision with front impact has 10,251 records; this is 98.70% of the 10,386 head-on collision records. We have therefore decided to limit our considerations to the cases where the initial point of impact was described as front impact. In the dataset used in our experiments a total of 67.68% of the records contained no information about the travel speed at the moment of impact. While other research indicates that the travel speed and speed limit are significant to the severity of injury resulting from an automobile accident, we decided not to utilize it as an input variable since there were too many unknown values. The final list of variables used in our experiments was: *drivers' age, gender, alcohol usage, restraint system, eject, vehicle body type, vehicle role, vehicle age, rollover, road*

*surface condition*, and *light condition*. The final dataset used for modeling had 10,247 records. In our research we have decided to approach the modeling problem by applying a one-against-all approach method. In this approach we select one output class to be *positive*, and all the other classes combined together to be the *negative* class. Then we run a series of five experiments with each class becoming the positive class in one of them.

## 3. EXPERIMENTAL SETUP, ANALYSIS AND RESULTS

We trained the neural network using a combination of backpropagation (BP) – 100 epochs, learning rate 0.01 – and conjugate gradient descent (CG) – 500 epochs – methods, trying to minimize the mean squared error. For each output class, we experimented with different number of hidden neurons, and report results obtained by the model with highest classification accuracy for the class. Average test performance values are reported in Table 1. Using a decision tree approach, we trained each class with the Gini goodness of fit measure. The prior class probabilities were set to equal, the stopping option for pruning was misclassification error, the minimum *n* per node was set to 5, fraction of objects was 0.05, the maximum number of nodes was 1000, the maximum number of level in tree was 32 and the number of surrogates was set at 5 (which turned to be the best decision tree parameters for the problem). The experimental results for the different classes are also summarized in Table 1.

**Table 1:** Test results for the different approaches

| Injury Class | ANN | | DT Accuracy (%) |
|---|---|---|---|
| | # Hidden Neurons | Accuracy (%) | |
| No Injury: | 65 | 60.45 | 67.54 |
| Possible Injury: | 65 | 57.58 | 64.40 |
| Non-incapacitating Injury: | 75 | 56.8 | 60.37 |
| Incapacitating Injury: | 65 | 61.32 | 71.38 |
| Fatal Injury: | 42 | 75.51 | 89.46 |

A number of observations can be made. First, the direct decision tree-based approach outperforms the direct NN approach in all cases. Second, the smallest difference is for the *non-incapacitating injury* category (4%), while the largest difference is for the *fatal injury* category (14%). Third, while such a comparison has only a limited validity, our results are better overall than these reported in [2] and indicate that decision trees are a serious contender to mining automobile accident data.

## 4. CONCLUSIONS

In this paper, we studied the National Automotive Sampling System General Estimates System automobile accident data from 1995 to 2000 and investigated the performance of neural networks and decision trees applied to predict drivers' injury severity in head-on front impact point collisions. As evident from Table 1, for all the five injury classes the decision tree approach outperformed neural networks. Previous research focused mainly on a binary classification of collision results into no injury and injury (including fatality) classes. In this paper, we were able to extend the research to include new categories: possible injury, non-incapacitating injury, incapacitating injury, and fatal injury. Furthermore, our experiments (details are omitted due to the lack of space) showed that the most important factors in fatal injury are: driver's seat belt usage, light condition of the roadway, and driver's alcohol usage. Finally, our experiments also showed that the model for fatal and non-fatal injury performed better than other classes. The ability of predicting fatal and non-fatal injury is very important since drivers' fatality has the highest cost to society economically and socially.

It has to be stressed again that it is a well-known fact that one of the most important factor differentiating injury level is the actual speed that the vehicle was going when the accident happened. Our dataset doesn't provide enough information on the actual speed, since speed for 67.68% of the data records' was unknown. If the speed was available, it might have helped to improve the performance of the two considered models.